\documentclass{article}
\usepackage{graphicx} 

\title{ICL Optimized Fragility}
\author{Serena Gomez Wannaz}
\date{September 2025}
\usepackage{float}
\begin{document}

\maketitle

\section{Abstract}ICL guides are known to improve task-specific performance, but their impact on cross-domain cognitive abilities remains unexplored. This study examines how ICL guides affect reasoning across different knowledge domains using six variants of the GPT-OSS:20b model: one baseline model and five ICL configurations (simple, chain-of-thought, random, appended text, and symbolic language). The models were subjected to 840 tests spanning general knowledge questions, logic riddles, and a mathematical olympiad problem. Statistical analysis (ANOVA) revealed significant behavioral modifications ($p < 0.001$) across ICL variants, demonstrating a phenomenon termed "optimized fragility." ICL models achieved 91\%-99\% accuracy on general knowledge tasks while showing degraded performance on complex reasoning problems, with accuracy dropping to 10-43\% on riddles compared to 43\% for the baseline model. Notably, no significant differences emerged on the olympiad problem (p=0.2173), suggesting that complex mathematical reasoning remains unaffected by ICL optimization. These findings indicate that ICL guides create systematic trade-offs between efficiency and reasoning flexibility, with important implications for LLM deployment and AI safety.

\section{Introduction: An Analysis of Optimized Fragility}

Building on the findings of a previous investigation on the effect of symbolic language on LLM reasoning, a broader research initiative has emerged. The first phase demonstrated that introducing an in-context learning (ICL) guide can modulate a model's behavior, making it faster but also more prone to errors. In that research, when presented with a fundamental variation in a riddle, the model consistently defaulted to the original solution, ignoring the new details. This behavior is notably similar to the \textit{A-not-B phenomenon} described in research by Han et al. (2024), where models adhere to a pre-existing pattern from examples, even when circumstances render it invalid.

The second phase of the research aims to delve into this phenomenon of optimized fragility by answering three key questions:

1.  \textbf{Does ICL affect other areas of knowledge?} While the initial study focused on logic and mathematics, it is crucial to determine if the impact of the ICL guide extends to other domains, such as history, creativity, and factual recall. Would the A-not-B phenomenon or a variant of it recur in these areas?
2.  \textbf{Is symbolic language the cause of the modification?} This question is to determine if "optimized fragility" is an exclusive effect of a proprietary symbolic language guide or if it is an inherent characteristic of introducing any ICL guide.
3.  \textbf{Do ICL variants modify the model's behavior?} I propose to investigate whether the model's original logic and processing are altered by introducing different types of guides, such as Chain-of-Thought (CoT) or randomized examples.

To explore these questions,  a more extensive experiment was designed with six models, including the original baseline model and five variants with different types of ICL. Each model was subjected to 140 tests across a variety of tasks, and metrics such as response time, median, and average were analyzed to detect and graph any changes in their processing and logic.

---

\section{Related Work}

\noindent
\textbf{In-Context Learning (ICL)} has emerged as a low-cost learning paradigm compared to traditional fine-tuning, eliminating the need for large training datasets for specific tasks. As demonstrated conclusively by \cite{Brown2020}, large-scale models can learn to perform tasks from just a few examples in a prompt. Other studies have highlighted ICL's ability to modulate a model’s behavior, enabling the adjustment of its personality or a specific role in domains such as customer service \cite{Handa2025}. These initial benefits positioned ICL as an agile and versatile method for model adaptation.

\noindent
However, despite these advantages, the scientific community has identified inherent vulnerabilities in LLMs. Research by \cite{Weber2025} has pointed out a lack of \textbf{consistency} and \textbf{non-robust generalization}, showing that model predictions can change unpredictably with minor prompt variations. This behavior reveals a fundamental fragility also manifested as a \textbf{degradation in task performance} \cite{Handa2025}. These findings suggest that optimizing for one objective often comes at a cost to a model’s versatility or accuracy.

\noindent
Expanding upon this line of research, the \textbf{A-not-B phenomenon}, observed in language models by \cite{Han2024}, exemplifies a \textbf{perseverance bias} where models adhere to a previous pattern even when circumstances render it invalid.

\noindent
While these problems have been identified, there remains a gap in understanding how introducing an ICL guide systematically \textbf{modulates} these vulnerabilities. This investigation addresses this gap directly by exploring whether different types of ICL transform a model's vulnerability, shifting its state from \textbf{"chaotic resilience"} to \textbf{"optimized fragility"} in tasks unrelated to the guide itself.

---

\section{Methodology}

To explore the impact of ICL on model fragility, a comparative experiment was designed using a reference model and a series of variants with different types of ICL. The experiments were conducted on the \textbf{GPT-OSS:20b} model. The only variable modified in each iteration was the Modelfile, which allowed the effect of the guide on model performance to be isolated. All prompts and responses were given in Spanish, and the test was automated with Python scripts to ensure consistency and efficiency in data collection. To validate the statistical significance of the differences observed between the groups, a \textbf{one-way analysis of variance (ANOVA)} was performed on the response times of the six models for each of the three test categories.

A critical aspect of the methodology is that the ICL guides were not directly related to the test questions. For example, in the case of the guide with symbolic language, the examples did not contain information about history or logic. This was designed to prove that ICL does not behave like a database or a training dataset, but rather as a catalyst for a new reasoning strategy, which validates the finding that In-Context Learning causes changes in the model's reasoning beyond simple data memorization.

\subsection{Test Groups}

Six model groups were tested with the following prompt configurations:

\begin{enumerate}
    \item \textbf{Base Model (Original):} Without any guide or ICL. This group served as the baseline for performance comparison.
    \item \textbf{ICL (Simple):} A prompt with direct examples of correct questions and answers.
    \item \textbf{ICL (CoT - Chain-of-Thought):} A prompt that includes intermediate reasoning steps to guide the model's response.
    \item \textbf{ICL (Random):} A prompt containing a series of semantically incoherent examples---lacking a logical pattern or relationship between inputs and outputs.
    \item \textbf{ICL (Appended Text):} A prompt with irrelevant "noise" text at the beginning, followed by ICL examples, to test for positional bias.
    \item \textbf{ICL (with Symbolic Language):} A prompt that uses symbolic language to guide the model, which was the focus of the previous report.
\end{enumerate}

\subsection{Dataset}

To evaluate each model exhaustively, a dataset of 14 questions was used, with 10 tests per question, for a total of \textbf{140 tests per model} and a total of \textbf{840 tests} throughout the experiment. The dataset included a variety of problems from different knowledge domains:

\begin{itemize}
    \item \textbf{10 general knowledge questions:} Including history, simple mathematical reasoning, and creativity. These questions were designed to evaluate the model's general capacity and reliability.
    \item \textbf{3 riddles with subtle variations:} For an empirical validation of "optimized fragility," three riddles from the initial report were reintroduced.
    \item \textbf{1 Math Olympiad problem:} This complex problem was used to evaluate each model's capacity for deep reasoning and resilience in high-difficulty tasks.
\end{itemize}

\subsubsection{The complete set of general knowledge questions was as follows:}

\textbf{Logic and Reasoning Questions}
\begin{itemize}
    \item \textbf{Math:} "If the statement 'All programmers are mathematicians' is false, which of the following statements must be true? a) No programmer is a mathematician. b) At least one programmer is not a mathematician. c) All mathematicians are programmers. d) No mathematician is a programmer."
    \item \textbf{Candy:} "There are 10 candies in a box. 6 are strawberry and 4 are lemon. If you take 2 candies at random, what is the probability that both are strawberry?"
\end{itemize}

\textbf{History and Facts Questions}
\begin{itemize}
    \item \textbf{Moon:} "On July 20, 1969, Neil Armstrong became the first man to walk on the Moon. What was his famous quote at that moment?"
    \item \textbf{Roman:} "Who was the first emperor of the Roman Empire and what name did he adopt upon assuming power?"
    \item \textbf{Medusa:} "In Greek mythology, which hero killed Medusa and how did he do it?"
\end{itemize}

\textbf{Math and Calculations Questions}
\begin{itemize}
    \item \textbf{Eggs:} "If a dozen eggs costs \$3.00, how much do 4 eggs cost?"
    \item \textbf{Walk:} "A person walks at 5 kilometers per hour. If they go from city A to city B, which is 20 kilometers away, and take a 30-minute break, how long will it take them to arrive?"
    \item \textbf{Consecutive Numbers:} "The sum of three consecutive integers is 36. What is the smallest of the three numbers?"
\end{itemize}

\textbf{Creativity Questions}
\begin{itemize}
    \item \textbf{Robot:} "Write a story about a robot that has to walk through the desert."
    \item \textbf{New Word:} "Create a new word and a definition for it."
\end{itemize}

\subsubsection{Logical Riddles}

For the research, three logical riddles with subtle variations were selected, designed to test the models' adaptability beyond the memorization of common answers. These riddles and their variations were based on previous work by Williams and Huckle \cite{Williams2024}.

\begin{itemize}
    \item \textbf{The Farmer's Riddle:} English: A farmer wants to cross a river and take with him a wolf, a goat, and a cabbage. He has a boat with three separate and secure compartments. If the wolf and the goat are alone on a bank, the wolf will eat the goat. If the goat and the cabbage are alone on the bank, the goat will eat the cabbage. How can the farmer effectively cross the river with the wolf, the goat, and the cabbage without anything being eaten?
    
    \item \textbf{The Robot's Riddle:} English: A robot has eight arms. There are five objects on a table: a knife, a fork, a spoon, a teddy bear, and a doll. The robot picks up each object with one arm. Then it shakes its own hand.
    
    \item \textbf{The Doors Riddle (Monty Hall Problem variant):} English: Imagine you're on a game show and are given the choice of three doors: behind one is a gold bar; behind the others, rotten vegetables. You choose a door, say number 1, and the host asks you, "Would you rather have door number 2?". Is it to your advantage to switch your choice?
\end{itemize}

\subsubsection{The Mathematical Olympiad Problem}

An \textbf{International Mathematical Olympiad (IMO)} problem was used to evaluate the models' capacity for deep reasoning and complex problem-solving. The problem, which is not solvable through pattern searching or memorization of solutions, was:

"A point $P$ is in the interior of a square $ABCD$ such that $PA = 1$, $PB = 2$ and $PC = 3$. Find the measure of the angle $\angle APB$ in degrees."

This makes it an ideal test to determine if ICL guides affect the model's ability to perform complex deductive reasoning, and if the models can resist the temptation to "fabricate" a solution when they do not have an answer in their training data.

\subsection{Evaluation Metrics}

Each model's performance was evaluated on a series of reasoning tasks to measure accuracy and efficiency. Responses were classified as follows:

\begin{itemize}
    \item \textbf{Score 2 (Correct):} Assigned to responses that were completely correct and did not contain errors or fabricated information.
    \item \textbf{Score 1 (Partially Correct):} Assigned to responses that were mostly correct but contained incorrect, fabricated, or hallucinated information. This type of response was more common in the original model's tests, indicating that its exploratory reasoning---part of its chaotic resilience---often sacrificed accuracy for breadth, leading to coherence errors or data fabrications.
    \item \textbf{Score 0 (Incorrect):} Assigned to responses that were completely incorrect and did not solve the problem or answer the question.
\end{itemize}
---

\section{Results}

The experiment revealed a consistent and significant difference in the models' behavior, validating the hypothesis that ICL guides, regardless of their content, induce a modulation in the model's reasoning strategy. The findings were grouped into categories for a clearer analysis.

\subsection{Analysis of Models in General Knowledge Questions}

The analysis of the 10 general knowledge questions showed a clear trend in most ICL-guided models, in contrast to the original model. ICL-guided models, in general, proved to be more efficient and reliable in these tasks. In addition to performance, response time measurements revealed a fundamental difference in processing speed and consistency.

\subsection{Mean, Median, and Standard Deviation}

The original model exhibited a clear \textbf{chaotic resilience}, manifested by high variability in its response times. Its average response time (\textbf{mean}) was \textbf{32.01 seconds}, while its median was \textbf{20.47 seconds}. This notable difference between the mean and median suggests the presence of atypically slow responses, a sign of its unpredictable and exploratory reasoning. The standard deviation of \textbf{30.83 seconds} underscores this high variability.

In contrast, the ICL-guided models showed a more optimized and predictable behavior:

* \textbf{ICL (Simple):} Mean of \textbf{13.33s} and median of \textbf{9.55s}, with a standard deviation of \textbf{9.45s}.
* \textbf{ICL (CoT):} Mean of \textbf{24.57s} and median of \textbf{19.73s}, with a standard deviation of \textbf{14.13s}.
* \textbf{ICL (Random):} Mean of \textbf{24.87s} and median of \textbf{16.04s}, with a standard deviation of \textbf{25.20s}.
* \textbf{ICL (Appended Text):} Mean of \textbf{15.46s} and median of \textbf{10.87s}, with a standard deviation of \textbf{12.65s}.
* \textbf{ICL (Symbolic):} Mean of \textbf{18.94s} and median of \textbf{12.91s}, with a standard deviation of \textbf{14.51s}.

The significantly lower standard deviation of the ICL-guided models reveals that these guides induced a more consistent and optimized reasoning process, sacrificing exploration for a more direct approach.

To provide statistical evidence, a one-way ANOVA confirmed that the differences in response times between the models were \textbf{statistically significant} ($F(5,834)=12.8777, p < 0.001$). This result provides robust quantitative evidence that the ICL guides generated a real and measurable effect on the models' processing speed.

\subsection{The Visualization of Consistency}

As can be seen in the scatter plot, the original model's response times are much more dispersed compared to the ICL-guided groups, which cluster more densely. This offers a visual representation of the original model's inconsistency, where its exploratory thinking manifests in a wide range of response times. In contrast, the ICL-guided models show a predictability that can be seen directly in their more clustered data.

\begin{figure}[H]
    \centering
    \includegraphics[width=0.8\textwidth]{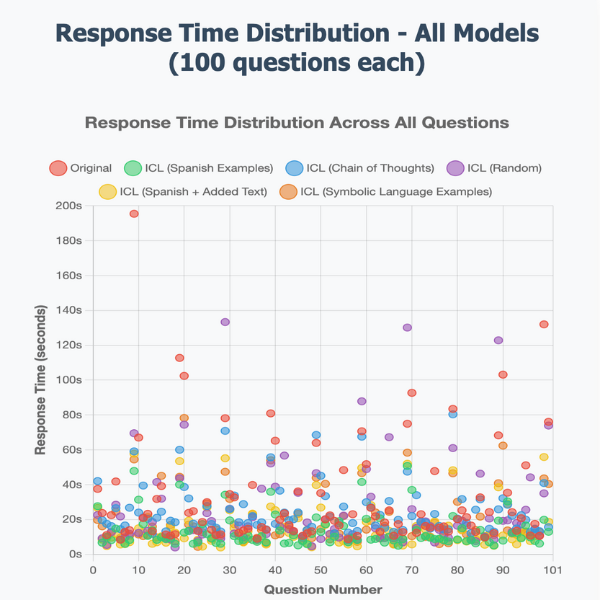}
    \caption{"Scatter Plot - Time Response General Questions".}
    \label{fig:resultados_generales}
\end{figure}

\subsection{Performance by Score}

The analysis of response scores revealed a crucial contrast between the original model and the ICL-guided models.

\textbf{Original Model (88\% accuracy):} This model had 88\% correct answers. An important finding, however, is that out of its 100 tests, 11 responses were partially correct (Score 1), and only one was completely incorrect. Additionally, the original model consistently produced more extensive and detailed responses than any of the ICL-guided models. This response pattern is a direct manifestation of its "chaotic resilience," where, lacking a guide, the model explores a wide range of options, often hallucinating or providing excessive information to fill gaps. The ICL (CoT) model came closest in terms of response length and detail.

\textbf{ICL Models:} In general, these models showed a more solid performance in terms of accuracy.
* \textbf{ICL (Simple):} 98\% accuracy (98 correct, 1 partial, 1 incorrect).
* \textbf{ICL (CoT):} 95\% accuracy (95 correct, 4 partial, 1 incorrect).
* \textbf{ICL (Random):} 93\% accuracy (93 correct, 4 partial, 3 incorrect).
* \textbf{ICL (Appended Text):} 99\% accuracy (99 correct, 1 partial).
* \textbf{ICL (Symbolic):} 91\% accuracy (91 correct, 6 partial, 3 incorrect).

The smaller number of partially correct responses in the ICL-guided models (with the exception of the symbolic language model) indicates that the guides helped them be more assertive, providing more precise answers or failing completely without attempting to guess.

\begin{figure}[H]
    \centering
    \includegraphics[width=0.8\textwidth]{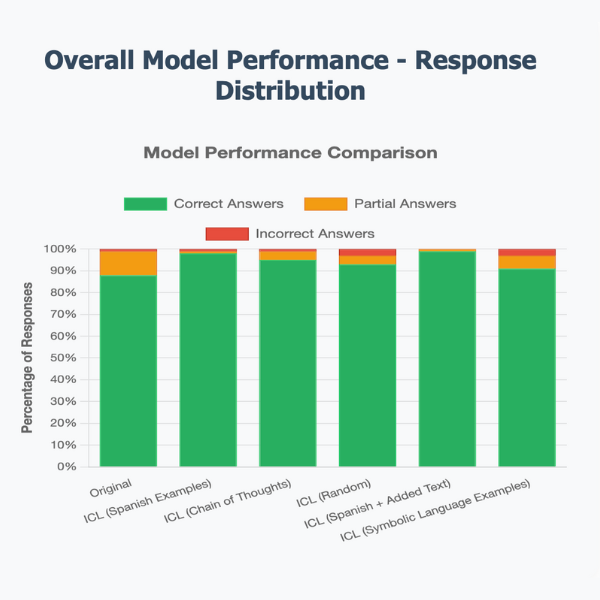}
    \caption{"Accuracy Responses General Questions".}
    \label{fig:resultados_generales}
\end{figure}

\subsection{Accuracy by Knowledge Area}

While ICL-guided models showed greater efficiency, the detailed analysis by question category revealed how each guide impacts the model's reasoning ability.

\textbf{Original Model's Performance (Chaotic Resilience):} The original model, without an ICL guide, exhibited inconsistent but effective behavior in certain areas. In the areas of reasoning and mathematics, the model achieved a 100\% accuracy. This result suggests that its exploratory reasoning, though slow, can be very effective in solving logical problems independently. However, this strength did not hold in other areas. In history, for example, the model had a notable drop, achieving only 10\% accuracy on the Medusa question. This pattern of uneven performance highlights the concept of "chaotic resilience."

\textbf{ICL Models' Performance (Optimized Fragility):} In contrast, the ICL-guided models showed a clear optimization in their performance.
* \textbf{ICL (Simple)} and \textbf{(Appended Text):} These models achieved 100\% accuracy in most areas.
* \textbf{ICL (CoT)} and \textbf{(Random):} These models also showed robust performance but had a notable drop on the Medusa question, achieving only 60\% accuracy.
* \textbf{ICL (Symbolic):} The symbolic language model, despite its high efficiency, showed a drop in its performance in areas like history, with 80\% accuracy on the Roman question and 30\% on the Medusa question. This validates the hypothesis that the guide optimized it for one task at the cost of its performance in other areas.

As can be seen in the bar chart, the impact of the ICL guides on accuracy is clear and varies by task type.

\begin{figure}[H]
    \centering
    \includegraphics[width=0.8\textwidth]{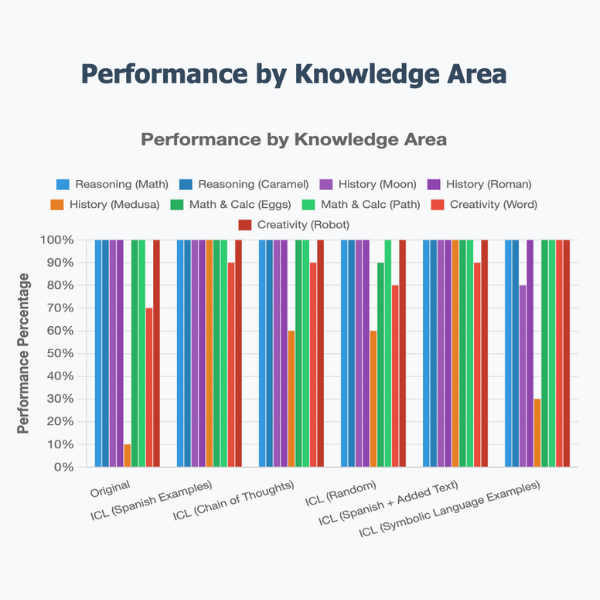}
    \caption{"Accuracy Responses By Knowledge Area".}
    \label{fig:resultados_accuracy}
\end{figure}

---

\section{Analysis of Models in Riddle Problems}

The original model continued to demonstrate chaotic resilience in the riddle tests, with high variability in its response times. Its average response time (\textbf{mean}) was \textbf{66.17 seconds}, while its median was \textbf{60.95 seconds}. This difference, along with a standard deviation of \textbf{33.15 seconds}, highlights its unpredictable and exploratory reasoning.

In contrast, the ICL-guided models showed a more optimized and predictable behavior:

* \textbf{ICL (Simple):} Mean of \textbf{15.15s} and median of \textbf{14.18s}, with a standard deviation of \textbf{5.32s}.
* \textbf{ICL (CoT):} Mean of \textbf{39.73s} and median of \textbf{31.1s}, with a standard deviation of \textbf{27.44s}.
* \textbf{ICL (Random):} Mean of \textbf{46.39s} and median of \textbf{32.4s}, with a standard deviation of \textbf{35.86s}.
* \textbf{ICL (Appended Text):} Mean of \textbf{25.42s} and median of \textbf{24.59s}, with a standard deviation of \textbf{15.75s}.
* \textbf{ICL (Symbolic):} Mean of \textbf{21.16s} and median of \textbf{19.16s}, with a standard deviation of \textbf{11.13s}.

The significantly lower standard deviation of most ICL-guided models demonstrates that these guides induced a more consistent and optimized reasoning process, sacrificing exploration for a more direct approach.

Similarly, a one-way ANOVA on the riddle response times confirmed that the differences between the models were also \textbf{statistically significant} ($F(5,174)=5.3294, p < 0.001$). This validates that the ICL guides had a real and measurable effect on the models' performance in this category, even when the effect led to a reduction in accuracy.

\begin{figure}[H]
    \centering
    \includegraphics[width=0.8\textwidth]{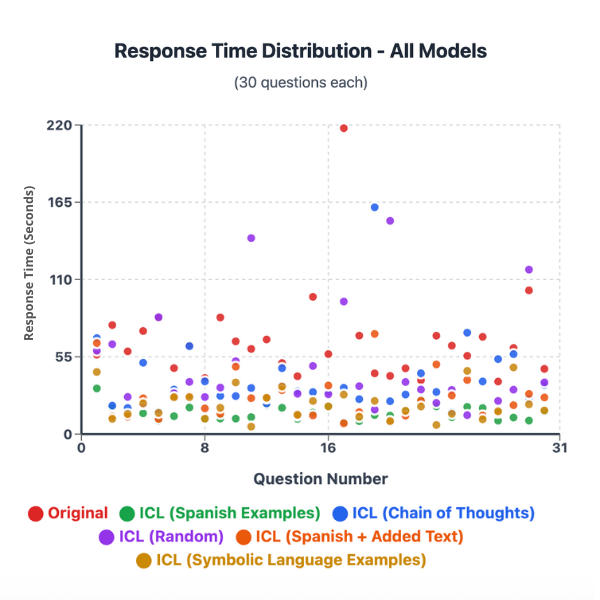}
    \caption{"Scatter plot - Time Responses".}
    \label{fig:resultados_scatter}
\end{figure}
\subsection{Accuracy and Model Behavior}

These results reveal the difference between presenting a model with an ICL guide with questions that have objective data or results and presenting it with three logic riddles that have variations. In the initial question tests, the ICL models showed greater effectiveness and correct answers. However, in the case of the riddles, this was not the case.

\subsection{Performance by Score}

The analysis of the riddle response scores revealed a significant contrast between the original model and the ICL-guided models.

* \textbf{Original Model:} It had 13 correct answers, 4 partially correct responses, and 3 incorrect responses, achieving an accuracy of 43\%. Its chaotic resilience was manifested in its ability to attempt the solution, often getting close to the correct answer (partially correct responses), but without a consistent approach.
* \textbf{ICL (Simple):} It got 5 correct answers, 5 partially correct responses, and 20 incorrect responses, resulting in an accuracy of 16\%.
* \textbf{ICL (CoT):} Unexpectedly, this model matched the original model's accuracy, with 13 correct answers, 0 partially correct responses, and 17 incorrect responses, reaching 43\% accuracy.
* \textbf{ICL (Random):} It showed low performance, with only 3 correct answers, 3 partially correct responses, and 24 incorrect responses, for an accuracy of 10\%.
* \textbf{ICL (Appended Text):} With 3 correct answers, 17 partially correct responses, and 4 incorrect responses, it achieved an accuracy of 30\%. The high number of partially correct responses is a notable finding, as the model attempted exploratory reasoning, similar to the original, but with a guide that oriented it toward a quicker result.
* \textbf{ICL (Symbolic):} It got 3 correct answers, 7 partially correct responses, and 20 incorrect responses, with an accuracy of 10\%. Like the other ICL models, it demonstrated a low capacity to solve the riddles, which is highly consistent with the optimized fragility that was discussed in the first report, related to a greater efficiency in time at the expense of a higher efficiency in responses.

These results show that while ICL optimizes models for knowledge tasks, its effect on complex reasoning varies. While the CoT model achieved an an accuracy comparable to the original model, most other ICL models performed significantly worse on the riddles.

\begin{figure}[H]
    \centering
    \includegraphics[width=0.8\textwidth]{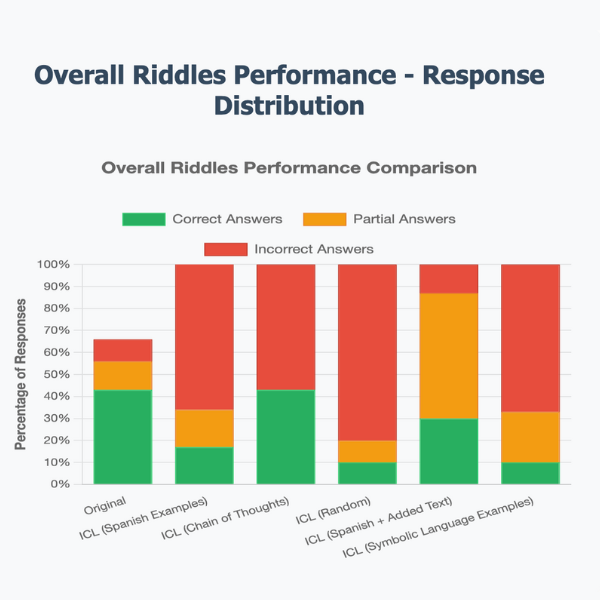}
    \caption{"Accuracy Responses Riddles".}
    \label{fig:resultados_generales}
\end{figure}
---

\section{Analysis of Models on Olympic Geometry Problems}

The Olympic geometry problem was selected to test the models' ability to perform complex deductive reasoning. Since it did not lend itself to memorization, it allowed for an impartial evaluation of the impact of each ICL guide. The results of the research suggest that ICL has a varied impact on each guide type, manifesting in different ways. It is worth noting that the ICL guides themselves did not contain any content related to solving these problems, nor any thought process logic, with the exception of the symbolic language guide in Model 6.

Notably, a one-way ANOVA on the response times showed no statistically significant difference between the models ($F(5,54)=1.4626, p=0.2173$). This result does not provide sufficient evidence to reject the null hypothesis, suggesting that if a difference exists, the experiment was not sufficiently powered to detect it. Further research with a larger sample size would be required to draw a definitive conclusion about the effect of ICL guides on deep reasoning problems.

\subsection{Mean, Median, and Standard Deviation}

1.  \textbf{Base Model (Original): Chaotic Resilience}
    The original model, without the influence of any ICL guide, behaved in the most unpredictable and chaotic way. It demonstrated chaotic resilience by being able to find correct solutions through exploration, but it also produced serious conceptual errors and hallucinations. Its response times were extremely erratic, with a \textbf{mean of 657.64 seconds} and a median of \textbf{439.41 seconds}, reflecting its exploratory behavior. Its time range, which varied from 232.18 to 1899.99 seconds, serves as the baseline for the research: a system that possesses flexibility but can generate an unwanted effect in more complex responses.
2.  \textbf{ICL (Simple)}
    This model, which was expected to be consistent and fast, showed that while most of its responses were grouped in an efficient range, it also produced a substantial failure that extended to \textbf{8402.79 seconds}. The difference between its \textbf{mean of 1703.25 seconds} and its median of \textbf{495.76 seconds} highlights how a single extreme failure can skew overall performance, indicating that optimized fragility can lead to a total collapse of reasoning.
3.  \textbf{ICL (CoT - Chain-of-Thought)}
    This model showed that its response times were dramatically inconsistent, with a \textbf{mean of 1108.89 seconds} and a median of \textbf{1104.38 seconds}. Although its solutions were structured in logical steps, they were often based on incorrect premises. The proximity between its mean and median indicates a consistency in its behavior, which suggests that the CoT guide forced it to follow a rigid logic, making the failure more predictable but also harder to detect.
4.  \textbf{ICL (Random)}
    This model exhibited the most erratic behavior of the ICL models. Its response times were equally erratic, with a \textbf{mean of 775.24 seconds} and a median of \textbf{839.12 seconds}. The closeness of these values, along with the high standard deviation, shows that an incoherent guide can be worse than having no guide at all, as the model cannot establish a reliable heuristic.
5.  \textbf{ICL (Appended Text)}
    This model demonstrated a vulnerability where, to produce a response, it attempted to apply a series of correct but often irrelevant theorems and propositions. Its response times varied from \textbf{216.00 to 1003.10 seconds}, with a \textbf{mean of 422.72 seconds} and a median of \textbf{349.10 seconds}. This manifestation of fragility highlights the dilemma between accuracy and efficiency, showing that a guide with "noise" text can result in overly complex responses.
6.  \textbf{ICL (with Symbolic Language)}
    This model presented a manifestation of optimized fragility: its responses, although often well-structured, were based on incorrect geometric premises and theorems. Its response times, with a \textbf{mean of 850.94 seconds} and a median of \textbf{682.62 seconds}, showed a more consistent behavior than the base model. This type of failure is insidious, as apparent formality can hide a lack of underlying understanding, generating convincing but false conclusions.

\begin{figure}[h!]
    \centering
    \includegraphics[width=0.8\textwidth]{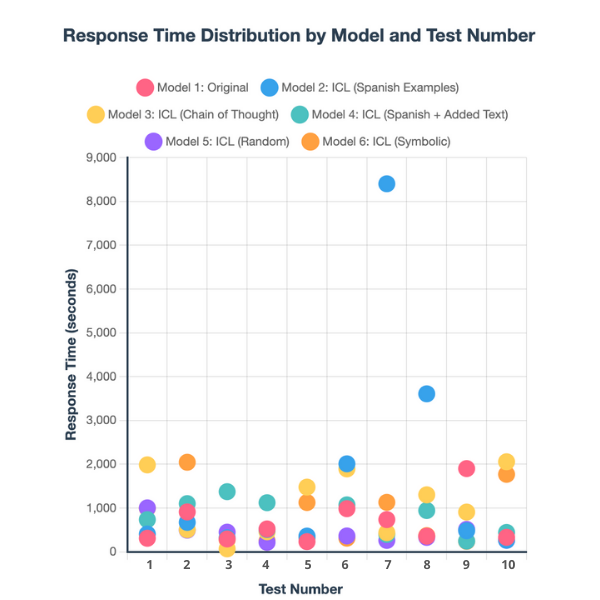}
    \caption{"Scatter plot - Math Olympics Problem".}
    \label{fig:resultados_generales}
\end{figure}
---

\section{Conclusions}

Through controlled experimentation and comparative analysis, the research supports the validation of the initial hypothesis that in-context learning (ICL) guides transform the inherent vulnerability of language models. The evidence suggests that ICL does not eliminate fragility; rather, it modulates it from a state of chaotic resilience to \textbf{optimized fragility}, the nuances of which manifest differently depending on the type of guide used.

\subsection{Answers to the Research Questions}

This phase of the research provides strong support for the following answers to the questions posed:

\textbf{Does ICL affect other areas of knowledge?} \textbf{Yes.} The results demonstrate that the impact of ICL transcends the domain of the knowledge presented in the examples. While the ICL models showed an improvement in efficiency and accuracy on general knowledge questions, their performance in complex reasoning tasks, such as riddles and geometry problems, was compromised, highlighting a cost to their knowledge transfer capacity.

\textbf{Is symbolic language the sole cause of the modification?} \textbf{No.} The findings suggest that the modification is an inherent characteristic of introducing any ICL guide. It was observed that each ICL variant (simple, CoT, random, appended text, and symbolic) affected the model's behavior uniquely, suggesting that the content and structure of the guide determine the type of heuristic shortcut the model adopts.

\textbf{Do ICL guides modify the model's behavior?} The evidence indicates that ICL fundamentally modifies the model's reasoning strategy. The response time and accuracy data demonstrate that the introduction of a guide is not a simple addition of data, but a catalyst that imposes a new logic of processing on the model.

\subsection{The Manifestation of Optimized Fragility}

The results by test category illustrate the dilemma of this fragility:

On general knowledge questions, the ICL models exhibited superior efficiency, achieving greater accuracy and consistency in their response times. This behavior indicates that the ICL guides optimized the models for direct information retrieval, promoting concise and direct answers instead of the detailed exploration seen in the original model.

On riddles and geometry problems, this optimization became a vulnerability. The original model, with its chaotic resilience, was able to explore diverse solutions, achieving more reliable performance than most of its ICL counterparts. The rigidity of the logic imposed by the ICL guides led the models to substantial failures or lower accuracy, as they became "trapped" in a procedure that was not suitable for reasoning that required creativity and adaptability.

\subsection{The Heuristic Shortcut as the Cause and Final Conclusion}

\textbf{Statistical Validation of Findings:} The ANOVA analysis provided robust evidence that the observed differences in model behavior were not anecdotal. The results on the general knowledge tests ($p < 0.001$) and riddles ($p < 0.001$) confirm that fragility modulation is a real and measurable phenomenon, while the lack of significance on the Olympiad problem (p=0.2173) reinforces the idea that the chaotic resilience of the base model is an advantage in high-complexity tasks.

The evidence suggests that each Modelfile served as a behavioral script that forced the model into a specific heuristic shortcut. The rigidity of the ICL (Simple), the step-by-step method of the CoT, the incoherence of the ICL (Random), and the focus on decoding of the ICL (Symbolic) were not merely data; they were procedures that redefined the way the model processes information.

Ultimately, this research suggests that, in the context of LLM security, optimizing for efficiency is not a neutral improvement but a compromise. ICL-guided models are faster and more predictable in data retrieval tasks, but in exchange, they sacrifice versatility and resilience in complex reasoning. This paradigm of \textbf{optimized fragility} represents a critical vulnerability that must be considered in the development and deployment of these systems.

The study provides quantitative and qualitative evidence that the introduction of ICL guides modifies the behavior of the GPT-OSS:20b model. The observed shift from chaotic resilience to optimized fragility is a significant finding, directly supported by the data.

A primary limitation of this work is its confinement to a single model architecture. This research serves as a foundational inquiry, and the study calls for further investigation into whether the modulation of fragility is a general characteristic of ICL across diverse LLM systems.

Finally, the most important question that arises from this work is how to design ICL guides that do not have a cost or a negative impact on the model. Further research could analyze whether these guides would create certain vulnerabilities or flanks within LLMs for possible exploitation.

---

\bibliographystyle{apalike}
\bibliography{references}

\end{document}